\documentclass[runningheads]{llncs}

\usepackage[T1]{fontenc}
\def\doi#1{\href{https://doi.org/\detokenize{#1}}{\url{https://doi.org/\detokenize{#1}}}}

\usepackage{amsmath}
\usepackage{xspace} 
\usepackage{amssymb}
\usepackage{graphicx}
\usepackage{comment}
\usepackage{algorithm}
\usepackage{todonotes}
\usepackage{subcaption}
\usepackage{eucal}

\usepackage{xcolor}

\definecolor{labelcolor}{cmyk}{0.22,0.10,0.10,0.10}
\definecolor{listbackgroundcolor}{cmyk}{0.10,0.10,0.05,0.05}
\definecolor{listbackgroundcolorlight}{rgb}{0.91,0.92,0.94}
\definecolor{candyapplered}{rgb}{1.0, 0.03, 0.0}

\usepackage{listings}
\begin{document}
\title{Towards Knowledge-Centric Process Mining}
%
%

\author{Asjad Khan\inst{1} \and
Arsal Huda\inst{2}\and
Aditya Ghose\inst{1}\and
Hoa Khanh Dam\inst{1}}
\authorrunning{Asjad K. et al.}
\institute{University of Wollongong,
Northfields Ave, Wollongong NSW 2522 \\
University of Nevada,
1664 N Virginia St, Reno, NV 89557\\
\email{asjad@outlook.com, syeda3@unlv.nevada.edu, aditya@uow.edu.au, hoa@uow.edu.au}\\
}

\maketitle              
\begin{abstract}


Process analytic approaches play a critical role in supporting the practice of business process management and continuous process improvement by leveraging process-related data to identify performance bottlenecks, extracting insights about reducing costs and  optimizing the utilization of available resources. Process analytic techniques often have to contend with  real-world settings where available logs are noisy or incomplete. In this paper we present an approach that permits process analytics techniques to deliver value in the face of noisy/incomplete event logs. Our approach leverages knowledge graphs to mitigate the effects of noise in event logs while supporting process analysts in understanding variability associated with event logs. 
Our approach is verified and validated on a sepsis event-log taken from a standard repository.




\end{abstract}
\section{Introduction}




Modern organizations routinely deploy process analytics, including process discovery and variant analysis techniques, both to gain insight into the reality of their operational processes and also to identify process improvement opportunities. Process analytic approaches play a critical role in supporting the practice of Business Process Management and continuous process improvement by leveraging process-related data to identify performance bottlenecks, reducing costs, extracting insights and optimizing the utilization of available resources. Over the past two decades, problems such as automated process discovery, process conformance checking and process enhancement have been extensively studied. Common to all process discovery approaches is the idea of extracting a process design from an event log which best represents the executions recorded in the process logs mined \cite{augusto2018automated}. Process mining techniques are primarily reliant on process logs, which don't always explicitly capture all the behaviour of past executed processes\cite{augusto2018automated}. Such logs are susceptible to domain gaps, data bias (due to incompleteness) and quality issues (due to noise and erroneous data recordings). Many real world processes are unstructured in nature and for these processes most state-of-the-art process discovery algorithms produce, hard-to-interpret, spaghetti-like models which poorly fit the event log. This negatively influences the usefulness of the discovered process model. Often times discovered models are hard to interpret from a process analyst perspective while also prone to under-fitting or over-fitting the given event logs, offering only minuscule support for improving process outcomes. Furthermore, a particular challenge in process mining is the management of business-process variants and contemporary business process management tools do not provide adequate support for modeling and management of process variants \cite{taymouri2021business}. For complex domains like healthcare, where improving clinical outcomes can directly impact the quality of life for patients, this implies that process analysts miss out on the opportunities for a complete understanding of the underlying process behaviour and subsequently extracting higher-impact insights. 

\textit{Common-sense reasoning} has been highlighted as one of the major challenges for the process analytics research community \cite{calvanese2021process}. It follows a broader trend in AI research where the need for solving complex tasks by  incorporating knowledge and \textit{common-sense reasoning} has been repeatedly highlighted \cite{davis2015commonsense}. In this work we investigate the challenges associated with process discovery and process variant analysis by considering a knowledge-graph based approach that assists process analysts in understanding the process execution behaviour from event logs. From a process analytics perspective, knowledge graphs offer rich semantics for knowledge representation and can be leveraged to model domain knowledge and process properties that are typically not captured in event logs \cite{gutierrez2021knowledge}. Knowledge graphs also provide structured relational knowledge between concepts, making them a good fit for tasks that require reasoning. These capabilities will  will allow process analysts to better understand process behaviour from event logs.   Our specific contributions are as follows: 



\textbf{Contributions:} We propose a knowledge-centric framework for supporting process analysts in their efforts to fully understand executed process behaviour recorded in real-world event logs. Our framework firstly aims to support the process discovery phase by leveraging knowledge graphs that enable constraint-based filtering of atypical behavior to improve the quality of mined dependency graph. Second, we propose a knowledge-graph based system that can facilitate the analysis of process variants from context-enriched event logs. Lastly, we tackle the problem of \textit{semantic incompleteness} of real-world event logs in order to improve the utility of extracted process models. We present preliminary evaluation on a real-world sepsis event-log representing sepsis cases in a hospital emergency room.

The paper is organized as follows: In Section 2 we provide the necessary background for understanding our proposed approach. We then present the motivation and details of our proposed framework in section 3. In section 4 we discuss related work followed by evaluation in section 5, where we discuss the evaluation setup, metrics used to evaluate our proposed machinery.

\section{Background and Related work}

\subsection{Process Mining} 

Processes deployed inside and across an enterprise, when executed can leave an operational data footprint in the form of an event logs, which if analyzed can be a valuable source of insights to support the management and improvement of business processes. This Process execution data can transformed into an event log describing the sequence of activities that were performed, along with the resources involved in the execution. Each entry of the process execution log(also known as event logs) represents an \emph{event}, which records the occurrence of an activity at a particular point in time and belongs to precisely one case (case represents a unique process instance). Events can be characterised by multiple descriptors (attributes), including an event (or activity name), a unique case identifier (e.g case ID), a timestamp and optionally details of resource responsible for executing the task of the process instance. An \textit{event} refers to an activity (or a step) in the process and belongs to a process instance or a case \cite{dumas2013business}. The sequence of ordered events within a case forms a trace.

\textbf{Definition 1 (Event Log) \cite{van2016process}: } An event $e$ is tuple $e=(c, a, t, r) \in$ $\mathcal{U}_{\text {case }} \times \mathcal{U}_{\text {act }} \times \mathcal{U}_{\text {time }} \times \mathcal{U}_{\text {res }}$ referring to case $c$, activity a, timestamp $t$, and resource $r$ of event e. An event $\log L$ is a multiset of events, i.e., $L \in \mathcal{B}\left(\mathcal{U}_{\text {case }} \times \mathcal{U}_{\text {act }} \times \mathcal{U}_{\text {time }} \times \mathcal{U}_{\text {res }}\right)$.

data as well). 

The process discovery phase involves analyzing business processes based on event logs (which are generated by most of today's information systems) to extract a process model and insights into the historical performance \cite{augusto2018automated}. Process Discovery algorithms can extract a business process model from an event log, which captures the control-flow relations between tasks recorded in the event log. Such models and insights can allow analysts to analyse performance and understand the behaviour of deployed processes. 

\subsection{Knowledge Graphs}

Over the past few years, Knowledge Graphs (KG) have emerged as a compelling abstraction for organizing enterprise knowledge \cite{gutierrez2021knowledge}. Knowledge Graphs  employ a graph-based data model to capture knowledge in a concise and intuitive abstraction for a wide variety of domains. In its simplest form, a knowledge graph is represented by a directed edge-labelled graph composed of nodes and edges. Nodes represent entities of interest and edges capture potentially complex relations between the entities of a given domain \cite{gutierrez2021knowledge}. Formally, given a set of nodes $N$, and a set of labels $L$, a knowledge graph is a subset of the cross product $N \times L \times N$. Each member of this set is referred to as a triple and labels capture meanings of the relationships between the entities represented by notes. 

Knowledge graphs are an effective tool for modeling interconnected, real-world scenarios while representing an ever-evolving substrate of knowledge within an organization. Knowledge graphs can be applied for integrating, managing and extracting value from diverse sources of data at large scale to present a unified view of enterprise knowledge \cite{ji2021survey}.  Knowledge acquisition tasks that are commonly performed to generate or extract implied information from Knowledge graph are knowledge graph completion, triple classification, entity recognition, and relation extraction \cite{ehrlinger2016towards}.




\section{Approach}



Process mining algorithm's performance is traditionally measured by how well it achieves pareto-optimality of the mined model in terms of various properties such as fitness and precision with respect to the available event log. The algorithm has an additional goal of achieving generalization on future process instances. In many practical settings, the target search space of models is quite large for an exhaustive search; therefore, process mining algorithms, enforce a specific representational bias to to make a trade-off (e.g. between higher fitness and lower precision).  In many real-world settings, process behaviour is not completely captured or available for mining in the event logs  \cite{augusto2018automated} as these logs are often \textit{noisy} and \textit{incomplete}. Process discovery algorithms when applied to real-world complex event logs often produce either noisy or incomprehensible models that either poorly fit the event log (low fitness) or over-generalize it (low precision or low generalization) \cite{augusto2018automated}. 

In machine learning research, recent interest in \textit{Neuro-symbolic AI} and \textit{Data-centric AI} techniques are based on the observations that we can make more theoretical progress in AI by reasoning about data instead of model architectures (e.g., number of layers or dimensions). Methods like data augmentation have gained popularity, allowing us to systematically engineer the data for building intelligent systems \cite{shorten2019survey}. In process mining, while a lot of emphasis has been on analyzing and extracting process insights from the observed behaviour logged in event logs, the knowledge dimension associated with business processes has received very little attention \cite{di2015knowledge}. i.e Current process mining techniques are self-contained and have minimal capacity to leverage and reasoning using prior knowledge. In practice, this means process mining algorithms can't perform inference that goes beyond the implicit knowledge which is recorded in the event logs. Traditional mining techniques focus on mining behaviour that inherently cannot represent all of the cascading hierarchical structure representing complex real-world processes \cite{van2009process}. These algorithms can't reason about abstract relationships between various objects involved in the process. For example, easily-drawn inferences that people can readily answer without direct training like \textit{smoke is seen so there must be fire happening} cannot be inferred by the current process mining methods. This leads to an incomplete understanding of process behaviour where process analysts are left trying to abstract, simplify, and even leave out key relationships needed for complete understanding of process behaviour.


Process knowledge has many faces and it will differ from other forms of organizational knowledge as it will be highly contextual, sometimes tacit and relevant to a particular domain. In this work we consider a knowledge-centric approach for assisting process analysts in understanding event data from multiple behavioral dimensions. Our framework addresses challenges associated with variability and data quality of event logs.  To explain our ideas we consider a real-world scenario where process mining techniques are applied. 

\subsubsection{Running Example - Sepsis Patient Administration:} At a high-level, medical diagnosis process consists of gathering data, classifying and diagnosing a specific problem and suggesting a particular course of treatment. The notion of a clinical process (often also referred to as a careflow) underpins the practice of medicine. Clinical pathways are care plans that attempt to standardise clinical or medical treatment processes. The process data representing such pathways, is often recorded in hospital information systems and clinical data warehouses. To illustrate the workings of our framework, let us consider an example of sepsis treatment careflow, where clinicians face complicated decision problems during the patient treatment process in an emergency room. In practice, sepsis treatment is highly knowledge-driven and consists of predictable and unpredictable elements. As the process evolves, knowledge workers (team of doctors) involved in the process make decisions based on clinical observations and patient data that is being constantly updated.



\begin{figure}[h!]
  \centering
  \begin{subfigure}[b]{0.9\linewidth}
    \includegraphics[width=\linewidth]{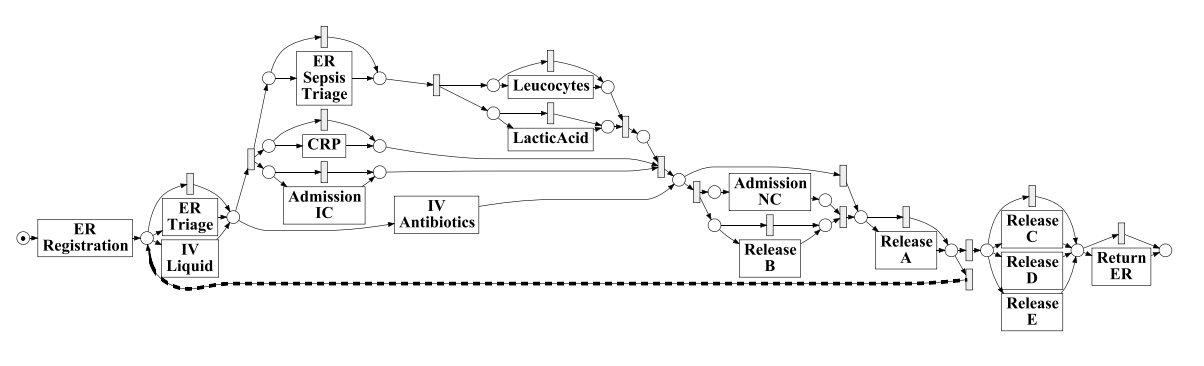}
    \caption{Process Model Mined using process discovery algorithm}
  \end{subfigure}
  \begin{subfigure}[b]{0.6\linewidth}
    \includegraphics[width=\linewidth]{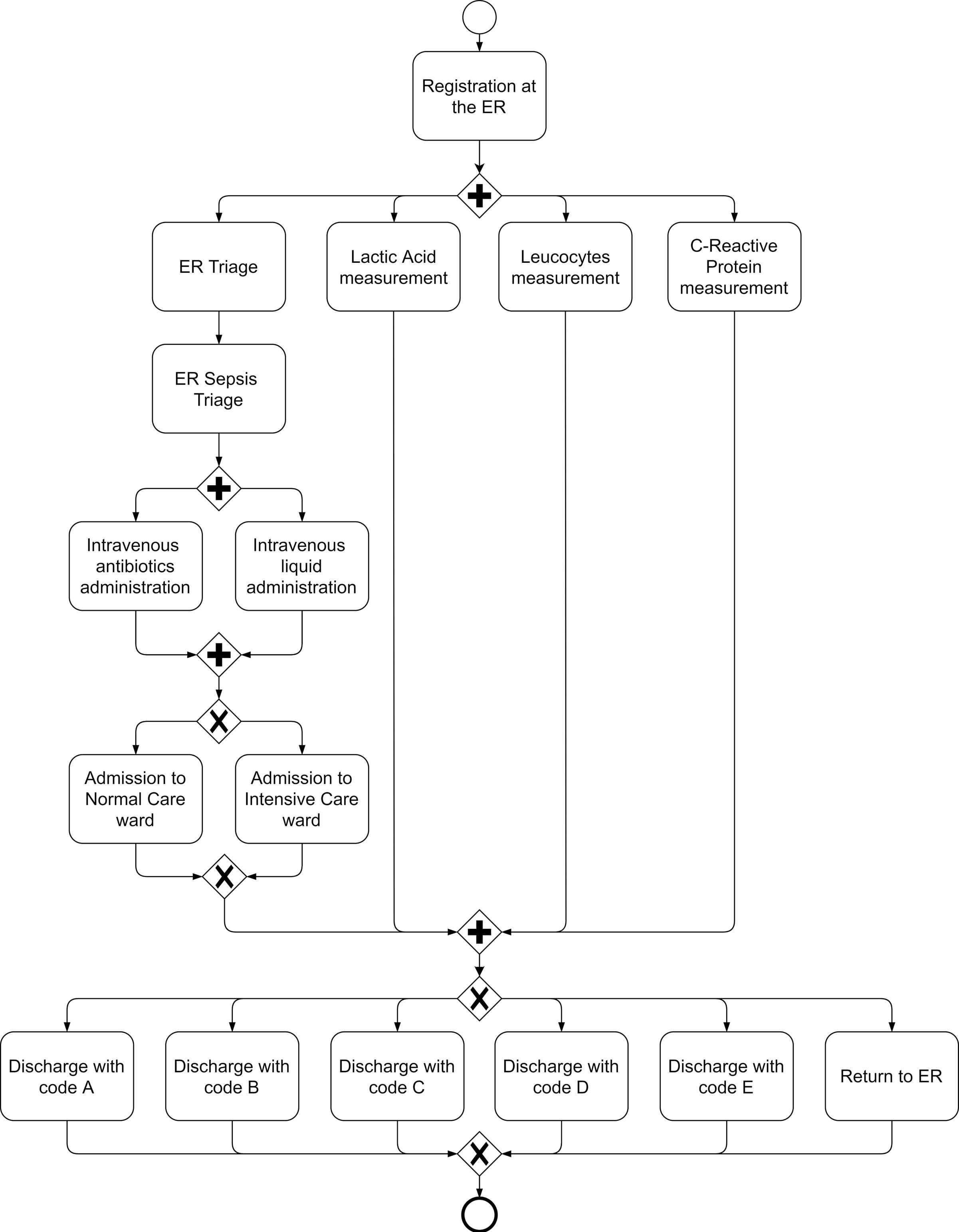}
    \caption{Actual Deployed Process Model}
  \end{subfigure}
  \caption{A comparison of mined vs expert drawn models}
  \label{fig:coffee}
\end{figure}

There is a growing recognition of the potential benefits of applying Process mining in understanding and optimizing the clinical care pathways of the patient treatment process. The vast array of personalized processes that are recorded in process histories, coupled with contextual information about patient demographics, medical history and co-morbidities can form a rich data source for mining insights. Process mining techniques assist process analysts in understanding the impact and correlation of clinical activities of a given patient.  We consider an event log, a real world example of sepsis treatment, representing treatment of 1050 patients over the course of 1.5 years, where events were recorded by the ERP (Enterprise Resource Planning) system of the hospital. There are about 1000 cases with total of 15,000 events that were recorded for 16 different activities \cite{mannhardt2017analyzing}. Here, we limit the complexity to keep the scope of our example manageable by focusing on a sub-group of patients and a subset of a mix of medical and logistical activities.  Event log typical is a good starting point for describing the sequence of activities that were performed. \\

The structural complexity, poor data quality and data incompleteness of available data is a major challenge during process discovery. Process mining algorithms can extract a process model as shown in Figure 1(a). However due to data quality challenges, some of the behaviour cannot be accurately extracted using the process discovery algorithms, limiting the mined model's utility. To understand these challenges we compare the mined model with an expert draw model which represents the on-ground executed process behaviour in Figure (B) \cite{munoz2022process}. Based on expert knowledge, we observe that even though IVs Administration and ER Triage can occur at the same time, the mined model represents them as a XOR (exclusive) connection. Similarly, the model restricts the repetition of labs test and doesn't provide the option for a patient Admission to Normal Ward before transferring them to Intensive care ward. Finally we observe that based on medical guidelines, time period between ER Sepsis Triage and IV Antibiotics should be less than 1 hour. We observe that  58.5\% of the cases are in violation of this rule, which indicates that the mined behaviour is mostly likely has been incorrectly recorded event data. This may happen as nurses may take a certain while to record the data as part of their administrative process. This example is indicative of the results of many real world process mining projects where often mining algorithms produce incomplete or Spaghetti (unstructured) models which require involvement of domain experts (e.g. subject matter experts) for a complete interpretation, limiting the broader applicability of process mining algorithms. 



We now present the details of key ideas underpinning our proposed machinery to address some of the above described challenges.

\subsection{Knowledge-based Dependency Graph Mining}

A core step in process discovery is mining a \emph{dependency graph} (also known as a {\em directly-follows graph}), which represents causal dependencies between tasks. The \textit{dependency graph} is a key abstraction of event data that is subsequently used to discover a process model. The dependency graph is defined as $D G=\{(a, b) \mid(a \in E \wedge b \in a \square) \vee(b \in E \wedge a \in \square b)\}$ Here $E$ is the finite set of activities, representing events that are recorded in the event $\log , \square b$ denotes the activities preceding $b$, and $a \square$ consists of the activities succeeding $a$ \cite{weijters2011flexible}. Computing the correct directly/eventually-follows relations from the event log can greatly improve the quality of the mined model. In this section, we investigate the application of knowledge graphs to enable constraint-based filtering of atypical behaviour during dependency graph mining. Automated filtering based on constraint checking (over domain knowledge and business rules) ensures that the dependency candidates generated by the process mining algorithm are correct and conform to a set of desired restrictions (e.g. defined by domain rules, policies etc). 


A knowledge graph (KG) offers a flexible way to conceptualise, represent and integrate process knowledge. From an organizational perspective, knowledge graph is useful for modeling explicit knowledge that typically represents process context, subject matters expertise of knowledge workers along with business rules and constraints. In this work, we assume the availability of a knowledge graph that encodes knowledge about the domain, business policies and regulations. In scenarios, where modeling a domain knowledge graph completely from scratch isn't feasible, we can use machine learning based methods for automated construction of a knowledge graph. Knowledge graph can be massively be scaled using natural language processing techniques, where we can build custom models that identify entities and relationships unique to a particular domain from unstructured text \cite{ji2021survey}. 


To rule out infeasible possibilities (and combinations thereof) in dependency relations, we consider a rule-based reasoning approach where we apply automated \textit{symbolic learning} to learn a model of logical formulae in the form of closed-path rules. Considering our sepsis example, these rules maybe be based on clinical guidelines developed by the medical community. In our sepsis example, the mined model incorrectly allows us to skip all activities between registration and discharge. A rule based on sepsis guidelines of the form \textit{For adults suspected of having sepsis, perform immediate sepsis screening test and admission during wait period} can constraint the model and prevent this from happening. Overall, rules provide a layer of conceptual knowledge needed to analyse event logs and to infer agreement/disagreement between \textit{observed} and \textit{true} behaviour and reach conclusions about unrealistic possibilities.  We note that true behaviour is not characterised by the designed process model rather by a set of all activities and their subsequent effects (post-conditions) that occurred on-ground during process execution. Automated rule discovery has a rich history and over the past few years, several scalable rule learners for knowledge graphs have been developed \cite{gutierrez2021knowledge}. In work we consider AMIE+, an efficient rule mining approach \cite{galarraga2015fast}, which allows us to mine closed-path rules(a language bias that is well-suited for rule learning in the context of knowledge graphs) in a self-supervised manner. To deal with an incompleteness aspect of the knowledge graph, we rely on Partial Completeness Assumption (PCA) and each rule has an associated confidence degree. 

\textbf{Definition 2 -  Closed-path rules \cite{omran2019embedding} } A CP rule (or simply a rule) $r$ is of the form
$$
P_{1}\left(x, z_{1}\right) \wedge P_{2}\left(z_{1}, z_{2}\right) \wedge \ldots \wedge P_{n}\left(z_{n-1}, y\right) \rightarrow P(x, y)
$$
Here $x, y$ and $z_{i}$ 's are variables, each $P(u, v)$ is called an atom, and $u$ and $v$ are called respectively, the subject and object argument for $P$. Intuitively the task of searching for CP rules can be reduced to that of finding plausible paths represented as sequences of predicates with high support and confidence. The derived rules capture the most representative constraints while considering the both structural and semantic features in the KG. 

To mine the dependency gtearaph from a given event log $W$, process discovery algorithms typically compute \textit{directly-follows} and \textit{eventually-follows} relations between two events a and b. For example flexible heuristic miner uses a frequency-based metric is used to capture the extent of dependency relation between two events A and B (represented by $a \Rightarrow_{W} b$) \cite{weijters2011flexible}.

$$a \Rightarrow_{W} b=\left(\frac{\left|a>_{W} b\right|-\left|b>_{W} a\right|}{\left|a>_{W} b\right|+\left|b>_{W} a\right|+1}\right)$$

The frequency metrics indicate the certainty level of a dependency relationship between two events $a$ and $b$ (high values indicate a strong relation) \cite{weijters2011flexible}. During this step we perform consistency checking in directly-follows relations against the mined rule-base. We rule out infeasible pairs if a directly-follows relationships is not entailed by the rule-base. Formally, for a given event log consisting of a set sequential event traces, constraint-base filtering involves finding pairs $(a, b)$ of sets of activities such that every element $\mathbf{a} \in \mathrm{A}$ and every element $\mathrm{b} \in \mathrm{B}$ are causally related (i.e., $a \rightarrow_{L} b$ ) and are consistent with each of the closed-path rules. This allows us ensure that the mining algorithm extracts the correct (feasible) event dependency patterns. Similar strategy is used for computing loops and long term dependencies.

\subsection{Knowledge-Aware Variant Analysis} 





Process execution traces exhibit a high degree of variability, representing deviating behaviour, especially in flexible environments such as the healthcare domain. In a medical context, variations can occur due to patients' and clinicians' agency, scientific and clinical evidence, and personal or organizational capacity \cite{sutherland2020unwarranted}. For treating sepsis patients,  deviations from clinical guidelines are fairly common in practice, to make allowance, for instance, for varying demographics (age, medical history) or for co-occurring conditions (sometimes, co-morbidities preclude the application of the standard protocol). Here, process owners are interested in identifying, quantifying, and reducing unwarranted clinical variation. To better understand the medical decisions underlying individual treatments, process analysts often carefully partition the data (care-flow instances) into cohorts based on the treatment provided by existing medical guidelines, patient profile (characterized by criteria like age group) and, the evidence base for prescribed treatment or patient's medical history.


Knowledge graphs have widely been applied for constructing context aware-based recommender systems where we use historic data to model item and user interactions. In our work, we construct a knowledge-graph base recommender system that encodes past variant behaviour as well as events annotated with contextual information to address the challenge of characterising deviating behavior. Our proposed machinery can assist the process analyst in classifying the traces based on the control-flow, shared cohort profile and contextual similarities and differences, associated with a given process instance. Formally, given an event log $L \subseteq \mathcal{E}^{*}$ as a set of traces $\sigma \in \mathcal{E}^{*}$, where each trace is  sequence of events annotated with relevant context and each trace is labelled with a variant class (as a training example) our machinery learns to partition the event log 
into a set of groups called process variants $\varsigma_{1}, \varsigma_{2}, \ldots, \varsigma_{n}$, such that $L=\varsigma_{1} \cup \varsigma_{2} \cup \ldots \varsigma_{n}$. Here process executions in the same group
must share the same contextual attribute value pair, and each process execution belongs only to one process variant.



In our sepsis example, we can consider mapping each input trace to three major patient cohort classes namely, \textit{effective care}, where trace variation implies the underuse of valid treatment or \textit{preference-sensitive care}, where variation implies availability of multiple care options and patient exercising choice), and \textit{supply-sensitive care,} where variation implies the volume of care provided is a reflection of organizational capacity rather than patient need \cite{sutherland2020unwarranted}.  We argue that process context mainly influences variant configuration and a wealth of relevant information can be found in non-process data (such as patient demographics and medical histories) which do not routinely appear in process logs, and which are not easily amenable to process mining. We consider event log augmentation which allows for a useful way to characterize the process context, enabling automated identification and classification of process variants. We therefore, annotate event logs  with contextual features (attributes) representing exogenous knowledge potentially relevant to the execution of the process that is available at the start of the execution of the process, and that is not impacted or modified during the execution of the process. We identify similar cases in logged data, that have been classified in the past and can form the basis for the current recommendation (annotated by process analyst as training examples). Each input trace is enriched by associating each event in the trace with a relevant attribute value pairs. Formally, given a event  $e=(c, a, t, r)$, and a context $C$ of a process instance, we augment the event with attribute-value pairs. 




 To understand variations, we consider labeled property graphs to encode behavioral and causal inter-dependencies of objects and actors over time in the context of process flows to symbolically represent a given situation for context-aware predictions. We model multi-dimensional event data as a labeled property graph in a manner similar to described in \cite{esser2021multi}. 

\textbf{Definition 4: (Labeled Property Graph: \cite{esser2021multi})} A labeled property graph (LPG) is a data structure where $K$ is a set of keys, $V$ is a set of values and Label be a set of labels. An LPG $G=(N, R$, label, prop ) consists of nodes $N$ (vertices) and relationships $R$ (edges) where each relationship $r \in R$ defines a directed edge $\vec{r}=\left(n_{1}, n_{2}\right)$ where $n_{1}, n_{2} \in N$.

The labeling function label : $N \cup R \rightarrow 2^{\text {Label }}$ assigns to each node and each relationship a non-empty set of labels designating their type, e.g., label $(n)=\{$ shock\_index, 0.5 $\}$. Function prop : $(N \cup R) \times K \rightarrow V$ assigns each node or relationship an arbitrary number of key-value pairs, called properties. For the value $\operatorname{prop}(x, k)=v$ of a property key $k$ of a node or relationship $x \in N \cup R$, we also write $x . k=v$ \cite{esser2021multi}.

Next, we consider embedding based models for scoring trace-class associations. For feature representation, we first generate the graph \textit{embeddings}, where we to fuse the trace-level and class-level representations to generate a knowledge-aware embedding vector.  We apply \textit{TransD} \cite{ji2015knowledge} method for entity representation learning.  This allows us to embed knowledge graph into a D-dimensional vector space using a distributional representation that captures semantic similarities among entities and predicates. Formally, given an input graph $G=(N, R$, label, prop ),  the encoding component maps entities to their distributed embedding representations. During training, the goal is to preserve local structure whereby vertices sharing similar edges in the graph are mapped to similar locations in the embedding. The distance between any pair vertices of as a measure of their relatedness and can be used to identify the presence or absence of a relationship. 

Next, we cast the recommendation problem as a link prediction problem where the model learns the high-order connectivity between an unseen event trace and associated class. We adapt the content-based model proposed by \cite{wang2018dkn} to learn a scoring function capable of generating class recommendations. Learned entity embeddings are taken as the input for a given an event $\log L$ we learn a prediction function $\widehat{y}_{\sigma \varsigma}=\mathcal{F}(\sigma, \varsigma \mid, \Theta, G)$ $\hat{y}_{\sigma i}$ which provides a relevancy score that predicts the probability that instance $\sigma_{n}$ belongs to particular cohort $\varsigma_{i}$. The model uses an attention module to learn high-order semantic relationship between each traces aggregated historical representation and associated variant class. During training it recursively propagates the embeddings from a node’s neighbors  to refine the node’s embedding, and employs the attention mechanism to discriminate the importance of the neighbors for providing accurate recommendations. Overall, 
the model allows context-aware characterization of process variants based on their cohort profile.

\subsection{Event log Augmentation}



Considering that the event log is the main input for process mining techniques, the quality of event logs is critical to the success of the process mining efforts. Process discovery from event logs remains a challenging problem as real-world process logs are often incomplete and don't explicitly capture all the behaviour of deployed processes. A domain expert might have a bunch of rules or facts about her domain, which can be useful for deducing more knowledge from the event log than what has been explicitly recorded. For example, reasoning on an available medical knowledge graph of infectious diseases along with patient data (like white blood cell count) will allow us to infer that patients who are compromised hosts(due immune problems) often also have gram-negative infections (caused by bugs like pesudo-monas). This kind of reasoning is challenging as; first, we require explicit specification of the conceptualisation of the given domain domain. Second reasoning mechanism must be robust to errors in the inputs to the system, to uncertainty in the knowledge, or to the gradual changes in the truth of various pieces of knowledge. 
The problem of Knowledge reasoning over knowledge graphs for inferring new conclusions from existing data has been well studied \cite{chen2020review}. Using this existing work as our foundation, we investigate the use of knowledge graphs for discovering missing events. Specifically, we address the \textit{semantic incompleteness} of an event trace, by inferring potential relations between entity pairs to automatically identify \textit{missing events}, based on existing knowledge with the purpose of complementing the event log. For a partially complete trace $\sigma$,over a event log $L$, our machinery generates a set of candidate  event traces $\sigma^{\prime}$ as a superset of $\sigma$ as output with the intent of improving the utility of mined models.

Similar to the approach taken in previous sections, we assume our knowledge graph encodes event data along with enterprise domain knowledge and business rules. To accurately capture the temporal patterns of event traces, we make use of a a temporal knowledge graph of the form $r(h, t \mid \tau)$ (where $h$ and $t$ represent entities, $\tau$ is a time stamp, and $r$ represents a relation that holds between these entities at the time specified by $\tau$). We adapt the technique proposed by Messner et al. \cite{messner2021temporal} where we use the box embedding approach to learn latent representations for events represented as entities, relations, and time stamps and then use the learned representations to predict missing events. We cast the event deduction problem as a temporal knowledge graph completion (TKGC) task where given the head (partially complete trace), we predict missing tail entity. Formally, our machinery computes the \textit{directly-follows degree} score representing the probability of missing successor event $e_{j}$ occurring after predecessor event $e_{i}$ $\left(e_{i} \rightarrow  e_{j}\right)$ in the trace $\left\langle\ldots, e_{i}, e_{j}, \ldots\right\rangle \in \mathcal{E}$. i.e Given the event log as premises along with general domain rules which are known apriori, we perform deductive reasoning to find new relations among graph entities in order to derive missing events. Overall, our machinery helps overcome the incompleteness aspect of event log and enables intelligent inferences from limited (or incomplete) amounts of recorded process data in order to guide process analysts better understand the process behaviour.

\section{Evaluation}




To evaluate our proposed framework, we use the same real-world sepsis data as described in our running example. We assume that the given sepsis log provides a partial view of the reality and that the expert designed model represents the ground truth. For comparison, we consider expert provided model to be a proxy of on-ground executed model (representing executed process behaviour). We then generate an augmented event log where we perform constraint based filtering followed by event log augmentation. Here, we address the \textit{correctness} challenge by filtering out chaotic activities against a rule-base extracted from a knowledge graph. Chaotic activities are deemed infeasible in the recorded event sequence of the process model based on domain and environmental constraints.  Second, we address \textit{incompleteness}, where we identify incomplete behaviour that is not explicitly recorded but is implied by the events recorded in event log. We quantify the utility of two different event logs by reporting precision (measures the ability of a model to generate only the behavior found in the log) and fitness (the extent to which the log traces can be associated with execution paths specified by the process model).  We use conformance checking operation to compare the behavior of the traces of the log against the footprint matrix of the expert designed model. A footprint matrix  describes for each couple of activities the footprint relationship and using Footprints conformance checking we investigate deviations and behavior of the log that is not allowed by the model. \textit{Fitness} and \textit{Precision} metrics are used to compare difference between `as-is' and `augmented' event logs produced by our machinery. Collectively, these metrics evaluate how well each trace in (augmented and raw) log fits the expert provided model and the degree of accuracy in which the process model describes the observed behavior.


\begin{table}[ht]
\centering
\begin{tabular}[t]{c c c c}
\hline
Event Log Type & Fitness & Precision & F-Score \\ [0.5ex] 
\hline\hline
Raw Sepsis log & 0.794 & 0.573 & 0.665  \\
Augmented  Sepsis log & 0.903 & 0.671 & 0.769  \\ [1ex] 
\hline
\end{tabular}
\caption{Framework Evaluation - A comparison of utility of two event logs}
\label{tab:caption}
\end{table}

We observe that our modified event log captures most of the behaviour described by the expert designed process model. This implies that process discovery algorithms should be able to accurately extract the behaviour represented in the process log (avoid overfitting). Viewed in this light, the empirical results suggest that the approach can indeed be useful in practical settings, in that our method improves the practical utility of available logs.

\section{Related Work}

Data quality issues associated with event logs, that may negatively impede model utility have received much attention over the last few years and a large and growing body of work has explored these challenges \cite{suriadi2017event}.  Event log improvement techniques have historically focused on issues like identification, visualization, and correction, or elimination of incorrect, or noise, missing, duplicate, and irrelevant events \cite{marin2021event}.  There is very little work available on combining knowledge with process analytic techniques to address the identified challenges. In a recent work Koschmider et al. \cite{koschmider2021demystifying} review the nature of data quality issues and also provide a summary of methods proposed thus far for detecting noise and outliers in event logs. They conclude that many of the noise filtering tools developed thus far do not appropriately filter noise.  Much of the literature on this topic addresses the problem by either pre-processing the logs before applying a process mining algorithm in order to correct the imperfect behaviors or using heuristic and patterns based mechanisms in conjunction with mining algorithms \cite{marin2021event} \cite{suriadi2017event}. In \cite{ayo2017probabilistic} Ayo et al. consider a  probabilistic approach to event log completenes. Their work relies on Fuzzy-Genetic Mining model based on Bayesian Scoring Functions for addressing the incompletness of event logs. Our proposed machinery on the other hand leverages knowledge graphs which encode externally available knowledge to address these challenges. In  \cite{taymouri2021business}  Taymouri et al. have performed a recent survey on techniques for variant analysis. For variant analysis, our work is similar to \textit{attribute
correlation} methods, where the goal is to cluster cases depending on event attributes. Closest to our approach is work by Leoni et al. \cite{de2016general} where decision trees with event attributed are considered to classify traces into process variants. However, our work differs from the above as we take a knowledge-centric approach where we take into account the process context in order to characterize variant behavior. 
In \cite{gudas2004framework} Dalmaris et al. have presented a framework based on process ontology. They advocate a point of view where process knowledge is considered the fundamental component of a comprehensive framework for business process improvement. They consider knowledge intensive processes and present in their framework a foundational theory of knowledge. i.e an ontology for dissecting, describing and discussing business processes and a method for the evaluating and improving of business process performance.


\section{Conclusion and Discussion}


In this work we investigate a knowledge-centric approach for addressing challenges associated with process discovery from noisy/incorrect event logs, context-aware variant analysis and semantic incompleteness of event logs. We argue that knowledge and processes are interlinked and knowledge should explicitly be made a key component of business processes mining practice. We show that our framework (in comparison with unaided classic process mining techniques) leads to improved utility and reliability of extracted process models. Overall, the ideas presented as part of our framework, provide a practical pathway for designing techniques that can tackle a wide range of other process analytic tasks. 



\bibliographystyle{splncs04}
\bibliography{references}

\end{document}